\documentclass[a4paper,fleqn]{cas-sc}

\usepackage[numbers]{natbib}

\def\tsc#1{\csdef{#1}{\textsc{\lowercase{#1}}\xspace}}
\tsc{WGM}
\tsc{QE}

\begin{document}
\let\WriteBookmarks\relax
\def\floatpagepagefraction{1}
\def\textpagefraction{.001}

\shorttitle{SeLIP: Similarity Enhanced Contrastive Language Image Pretraining for Multi-modal Head MRI}    

\shortauthors{Zhiyang Liu, Dong Yang, Minghao Zhang et al}  

\title [mode = title]{SeLIP: Similarity Enhanced Contrastive Language Image Pretraining for Multi-modal Head MRI}  



%

\author[1, 2]{Zhiyang Liu}


\credit{Conceptualization, methodology, funding acquisition, writing—original draft preparation, review, and editing }

\affiliation[1]{organization={College of Electronic Information and Optical Engineering, Nankai University},
            addressline={Tongyan Road}, 
            city={Tianjin},
            postcode={300350}, 
            state={Tianjin},
            country={China}}

\author[1]{Dong Yang}

\credit{Methodology, software, and writing—original draft preparation}

\author[1]{Minghao Zhang}

\credit{Software and writing—original draft preparation}

\author[1]{Hanyu Sun}

\credit{Investigation and software}

\author[1, 2]{Hong Wu}

\credit{Investigation and writing—review and editing}

\author[3]{Huiying Wang}

\credit{Data collection and annotation}

\author[3]{Wen Shen}

\credit{Data collection, annotation, and conceptualization}

\author[3]{Chao Chai}
\cormark[1]

\credit{Data collection, conceptualization, funding acquisition, and writing-review and editing}

\author[3]{Shuang Xia}
\cormark[1]

\credit{Data collection, conceptualization, funding acquisition, and writing-review and editing}

\affiliation[2]{organization={Tianjin Key Laboratory of Optoelectronic Sensor and Sensing Network Technology},
            addressline={Tongyan Road}, 
            city={Tianjin},
            postcode={300350}, 
            state={Tianjin},
            country={China}}

\affiliation[3]{organization={Department of Radiology, Imaging Medicine Institute of Tianjin, Tianjin First Central Hospital, School of Medicine, Nankai University},
            addressline={Fukang Road}, 
            city={Tianjin},
            postcode={300192}, 
            state={Tianjin},
            country={China}}

\cortext[1]{Corresponding author}




\begin{abstract}
Despite that deep learning (DL) methods have presented tremendous potential in many medical image analysis tasks, the practical applications of medical DL models are limited due to the lack of enough data samples with manual annotations. By noting that the clinical radiology examinations are associated with radiology reports that describe the images, we propose to develop a foundation model for multi-model head MRI by using contrastive learning on the images and the corresponding radiology findings. In particular, a contrastive learning framework is proposed, where a mixed syntax and semantic similarity matching metric is integrated to reduce the thirst of extreme large dataset in conventional contrastive learning framework. Our proposed similarity enhanced contrastive language image pretraining (SeLIP) is able to effectively extract more useful features. Experiments revealed that our proposed SeLIP performs well in many downstream tasks including image-text retrieval task, classification task, and image segmentation, which highlights the importance of considering the similarities among texts describing different images in developing medical image foundation models.
\end{abstract}

\begin{keywords}
 Deep learning\sep contrastive learning\sep image-text alignment\sep structural information extraction\sep multi-modal MRI
\end{keywords}

\maketitle

\section{Introduction}

Deep learning (DL) has been widely adopted in processing medical images \cite{li2023deep,rana2023machine,mall2023comprehensive,ronneberger2015u, hatamizadeh2022unetr,ali2022monkeypox,jiang2023deep, mijwil2023pixels,wang2020review}, and presented great potential in areas, such as tissue segmentation \cite{ronneberger2015u, hatamizadeh2022unetr}, lesion detection \cite{ali2022monkeypox}, medical diagnosis \cite{jiang2023deep, mijwil2023pixels}, and treatment planning \cite{wang2020review}. The DL models, however, are usually developed on specific datasets for some certain disease, and therefore the versatilities of the developed DL models are significantly limited\cite{shen2020introduction}. In practice, when developing a DL model for some disease or on images acquired from a new type of machine, one has to collect and annotate many images to train the model, even though a public dataset of the same task might be available. Despite that many few-shot learning methods have been developed based on a backbone network pretrained on large datasets such as ImageNet\cite{deng2009imagenet}, they are still difficult to be directly applied in medical images due to the lack of a well-trained backbone network with sufficient versatility.

The pretrain-finetune paradigm has been proven its effectiveness in natural language processing (NLP). A foundation model that trained on a huge dataset can be easily transferred to various downstream tasks by finetuning it using a limited number of samples. In the field of computer vision, mask autoencoders\cite{he2022masked} and self-supervised learning\cite{liu2021self, jaiswal2020survey}, have been extensively studied in pretraining a vision foundation model, and are able to achieve good performance in many downstream tasks. For instance, UNITER\cite{chen2020uniter} incorporated extra masked language modeling (MLM) and masked region modeling (MRM) tasks during the pretraining, following the ideas of BERT\cite{devlin2018bert}. A contrastive language-image pretraining (CLIP) method was also proposed to use contrastive learning between images and their corresponding text descriptions \cite{radford2021learning}. By pretraining the model on 400M image-text pairs, the model is able to achieve better performance on many datasets over conventional supervised learning methods. 

Recently, contrastive learning of medical images and texts has also been extensively studied. Different from the CLIP-like methods which focus on the ordinary images and can be trained on billions of image-text pairs, medical image-text pairs are more difficult to be collected, and therefore the main challenge becomes how to make better use of the limited number of data samples. It should be noted that most radiology examinations can naturally construct a dataset for visual language contrastive learning. 
The outcome of a clinical radiology examination includes the image and a radiology report. A report typically includes two sections, that is, ``\textit{Findings}'' and ``\textit{Impressions}'', where the ``\textit{Findings}'' section describes what radiologists see from the images, and the ``\textit{Impressions}'' section draws conclusions from the image descriptions. Such clinical protocol indicates that images, such as X-Ray, computer tomography (CT), and magnetic resonance imaging (MRI), will always be associated with texts which precisely describes the abnormalities and the appearances. This motivates us to develop a pretraining framework for medical images by using the paired images and reports. Compared with contrastive learning on natural image-text pairs where millions of samples can be obtained from the Internet, the number of available medical image-report pairs is much smaller. Some images, such as CT and MRI, are 3D images, which imposes more challenges.

The reports of single-modality images, such as X-ray and CT, are easier to use, as report texts are describing the same image. For examinations with multiple modalities, such as MRI, the report findings describe the symptoms observed from various modal images, making it difficult to train. As a result, it is necessary to first extract the human-written reports to structured reports, so that the descriptions of different modalities can be separated from each other. The clinical entities can be either extracted by training a classifier using BERT \cite{harnoune2021bert} or using the large language models (LLMs). For instance, \cite{meoni2023large} developed a clinical entity extraction method using LLM to avoid the risk of privacy leakage when using commercial LLMs. As we can see in this paper, the structured report can be well generated by finetuning a local LLM using pseudo-data. 

Conventionally, CLIP uses an identity matrix as the learning target of the cosine similarity matrix, so that the unpaired images and texts could be pushed away from each other. In the contrastive learning of medical image-report pairs, however, as pointed out in \cite{wang2022medclip}, even though some reports do not belong to the target patient, they can still describe the same symptoms and findings. Therefore, it is not reasonable to use the identity matrix as a hard target in MRI pretraining. Motivated by SoftCLIP \cite{gao2024softclip} which applied a soft semantic similarity target in addition to the hard CLIP target, in this paper, we proposed a soft similarity matrix that measures the similarity between texts from both the syntax and the semantic levels. The Kullback–Leibler divergence between the image-text similarity matrix and the soft similarity matrix is further adopted in addition to the conventional InfoNCE loss. We named the method as similarity enhanced contrastive language image pretraining (SeLIP). Experiments revealed that our proposed method is able to train the image and text encoders efficiently even when training on a small dataset with a small batch size. Our pretrained encoders are evaluated on multiple downstream tasks. The proposed method is able to present better performance over the comparative methods in all considered tasks, which highlights the importance of introducing soft targets in pretraining medical images.

Our main contributions are summarized as follows.
\begin{itemize}
    \item A contrastive learning framework for multi-modal MRI pretraining is proposed, and the language of texts is in Chinese and without any special tokens.
    \item A mixed syntax and semantic similarity matching method is applied to enhance performance, which alleviates the problems originated from the high similarities among radiology reports.
    \item The proposed SeLIP performs well on various downstream tasks, including image-text retrieval, image classification, and image segmentation, which highlights the importance of using semantic similarity as targets in training medical image foundation models. 
\end{itemize}

\section{Related Work}
\subsection{Vision Language Contrastive Learning}

In the image processing field, the large scale vision language models trained by contrastive learning have achieve notable achievements. The success of CLIP provided a novel way to train a foundation model in an annotation efficient way, followed by which researchers developed many vision models on much larger datasets \cite{jia2021scaling, pham2023combined, zhai2022lit, cherti2023reproducible}. ALIGN \cite{jia2021scaling} scaled up the number of training samples to 1.8B image-text pairs by relaxing the filter criteria in data collection. BASIC\cite{pham2023combined} further increased the dataset size to 6.6B and discussed the impacts of batch size, data, and model scale. LiT\cite{zhai2022lit} proposed to freeze the image encoder and train the text encoder on 18B samples, which simplified the task and achieved better performance on image-text alignment. 

Despite that better performance can be expected when using a large pretraining dataset, it is also important to investigate how to efficiently utilize the training samples when it is not possible to collect such a large amount of data. To this end, frameworks and models with additional supervision and information process have also been extensively studied. \cite{li2021supervision} proposed to use three intrinsic supervision to train the model, which outperforms CLIP based model while using $7.1\times$ fewer data. 
\cite{shi2025umg} constructed a new dataset with image-level, region-level and pixel-level pseudo captions and tag, and proposed to train the model on both global images with texts and local regions with captions. 
Another way to study is that how to enhance the performance of models by improving the training framework and methods of data utilization. 
\cite{cui2022contrastive} provided a reproducible baseline named ZeroVL which was with only publicly accessible academic datasets with 14M samples, mainly benefited from a debiased sampling method and a data augmentation method by coin flipping mixup. 
In \cite{fan2024improving}, researchers focused on the diversities of texts and proposed a language augmented CLIP models with language rewrites by LLMs. 

\subsection{Contrastive Learning on Medical Images}

While it seems that the vision language contrastive learning methods can be straightforwardly applied to medical field, there are also many additional challenges to tackle. Compared to natural images and texts, the datasets of medical images are typically much smaller, and the regions of interest on medical images are also small in most cases. Moreover, the texts that describe the images are usually lack of diversity. 
\cite{zhang2022contrastive} proposed a bidirectional method, named as ConVIRT, for contrastive learning between medical images and texts. In \cite{huang2021gloria}, GLoRIA, a pretraining framework with attention mechanism, was proposed, which conducted contrastive learning on both the global and local features and the texts. 
LRCLR\cite{rizvi2023local} proposed a contrastive learning method with flexible local region selection, which could better distinguish the subtle differences between medical images.
MLIP\cite{li2024mlip} introduced divergence encoders to enhance the generalization ability of the model, and provided the approach to integrate the token-knowledge-patch alignment and knowledge-guided prototype clustering contrastive learning. 

Radiology reports, although precisely describe the appearance of the medical images, the texts are usually lack of diversity compared to common texts, making it difficult to train, especially taking the limited number of medical image-text pairs into consideration.  
Many studies have shown that in the medical field, simply treating unpaired radiology reports as negative samples during the pretraining may have an adverse effect on model learning\cite{wang2022medclip, liu2023improving, chen2023knowledge}.
As a result, MedCLIP\cite{wang2022medclip} introduced a semantic matching loss that reflects the similarities among the texts to train the image encoder. 
\cite{liu2023improving} divided the image-text pairs into positive, neutral, and negative, and used the cosine similarity to evaluate the semantic similarity, in order to alleviate adverse effect of similar texts. 
\cite{liu2024towards} applied supervision on three perspective, namely cross-modal, multi-modal, and uni-modal, to uncover potential information within the medical images and corresponding texts. 
UMCL\cite{wang2023umcl}, by noticing the impact of false negatives, proposed to use softened image-text score matrix as target.

As pointed out in \cite{chen2023knowledge}, in medical vision language contrastive learning, the main challenges are semantic overlap and semantic shifting. The above mentioned literature \cite{wang2022medclip, zhang2022contrastive, huang2021gloria, rizvi2023local, liu2023improving, liu2024towards, wang2023umcl} mainly focused on solving the semantic overlap problem. To deal with the semantic shifting problem, text  standardization in different levels should be considered. MedKLIP \cite{wu2023medklip} proposed to first extract medical-related information from the raw reports by converting the reports as triplets of \textit{entity}, \textit{position}, and \textit{exist}, which filtered out the interference from grammar. 
KAD\cite{zhang2023knowledge} enhanced the performance of model by pretraining with a knowledge graph, where the clinical entities and relations are extracted from a reports information extraction toolbox or ChatGPT.  

Note that most articles \cite{wang2022medclip, zhang2022contrastive, huang2021gloria,rizvi2023local, liu2023improving, liu2024towards, wu2023medklip, zhang2023knowledge,wang2023umcl, li2024mlip} investigated the language image contrastive learning on chest X-ray images by using datasets such as MIMIC-CXR dataset\cite{johnson2019mimic} and CheXpert\cite{irvin2019chexpert}. As the X-ray images are in essence gray-scale and 2D ones, the contrastive learning methods developed for 2D RGB images can be applied in a straightforward manner. Such methods can hardly be applied to 3D and multi-modal images, such as MRI, as they contain more enriched information and require much more memory to train. In \cite{lei2023unibrain}, a framework called UniBrain was proposed for contrastive learning on multi-modal head MRI. It incorporated four image encoders to process four modalities, i.e., T1-weighted imaging (T1WI), T2-weighted imaging (T2WI), T2 fluid attenuated inversion recovery (T2FLAIR), and diffusion weighted imaging (DWI). In addition to CLIP head, a classifier head was also adopted as an auxiliary to train the network. While UniBrain provided an effective way to pretrain the image encoders for multi-modal 3D images, the design of four parallel input heads image modalities limited the cross-modal ability of the model, and the performance would also be degraded when there is one or more missing modalities.

\subsection{Contrastive Learning with Softened Targets}

Target softening aims to alleviate the constraints caused by hard target when different targets are non-independent, so as to reduce the model's overconfidence towards special cases to a certain extent. For example, in the field of knowledge distillation, the soft target of the student model is the logits predicted by the teacher model\cite{hinton2015distilling}, and learning the distribution and relationships between targets is more important. Recently, softened targets are introduced into contrastive learning method. SoftCLIP\cite{gao2024softclip} applied a softened target to achieve a soft cross-modal alignment. This research suggests that taking both the hard label contrastive learning and softened target constrained distributions in the training can achieve better performance. Pyramidclip\cite{gao2022pyramidclip} also softens the targets of negative samples to alleviate mistakes when the model is trying to distinguish compatible negative pairs. 
CLIP-PSD\cite{andonian2022robust} alleviated the influence when an image would be aligned to several text captions to different degrees. 
In MaskCon\cite{feng2023maskcon}, the soft labels based
on sample distances were applied to address a finer labelling problem. 
When applying contrastive learning to the medical field, the diversity of texts is greatly reduced, so the target needs to be softened appropriately.

Moreover, many methods of medical vision language contrastive learning which try to solve the semantic overlap problems are also applied by softened target in \cite{wang2022medclip, liu2023improving, chen2023knowledge, liu2024towards, wang2023umcl}.
Thus, the softened target is more suitable for correcting and supplementing the errors and insufficiencies of hard labels when processing medical image-text data.

\section{Materials and Method}
\label{sec:method}

\subsection{Data}

\begin{figure*}[!h]
\centerline{\includegraphics[width=.8\textwidth]{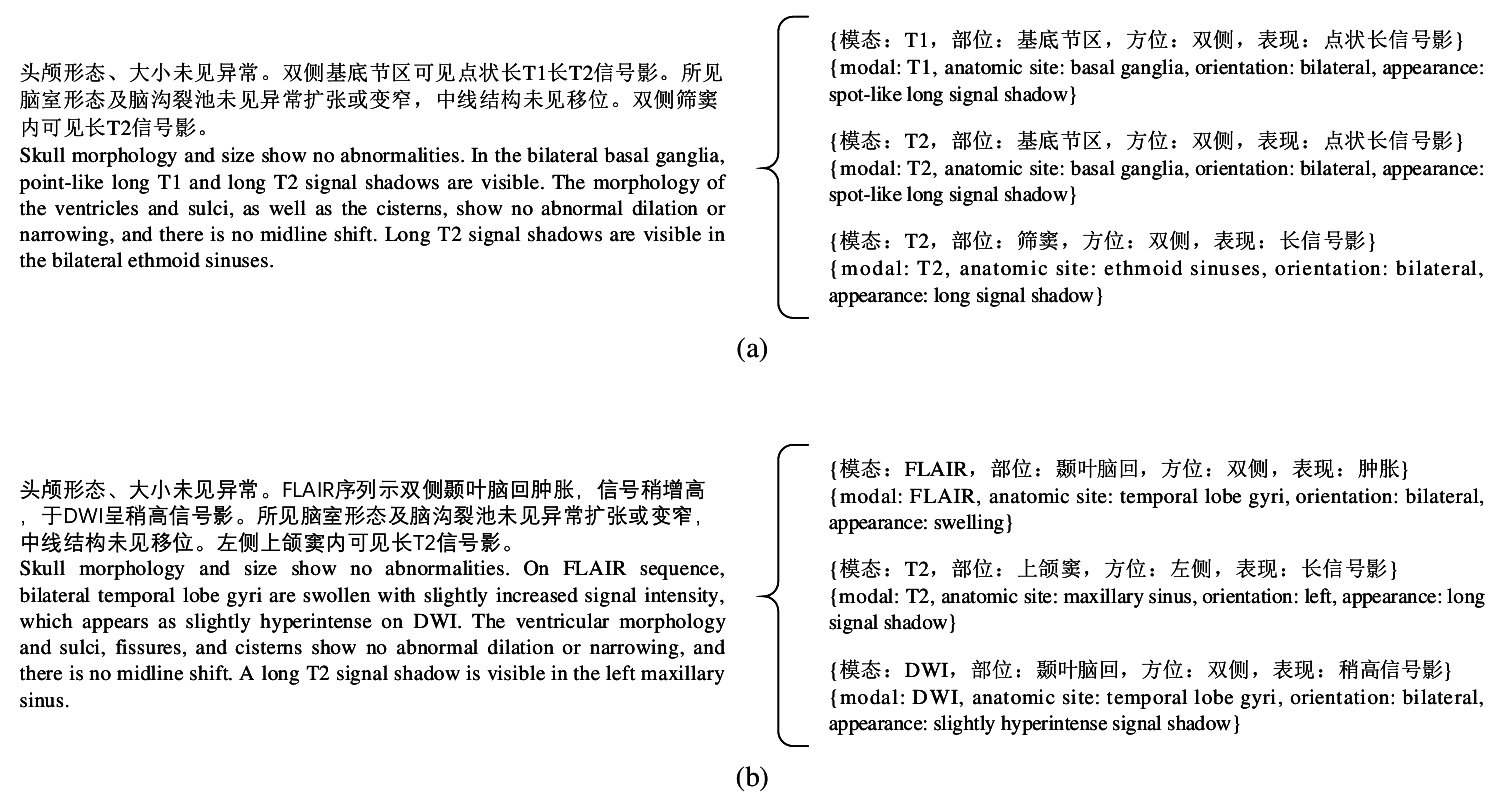}}
\caption{Examples of the ``Findings'' descriptions and the extracted structural information.}\label{fig:structure extraction}
\end{figure*}

We retrospectively collected head MRI images and the corresponding radiology reports of 10825 subjects from Tianjin First Central Hospital, Nankai University (Tianjin, China). The study was approved by Nankai University Institutional Review Board (NKUIRB2024001). Each subject may undertook MRI examination of various modalities as suggested by their clinicians, as summarized in Tab. \ref{tab: modalities}. The radiology report includes two sections, i.e., ``\textit{Findings}'' and ``\textit{Impressions}'', where the ``\textit{Findings}'' describes in detail about what the radiologists observe from the images, while the ``\textit{Impressions}'' draw conclusion on the radiology diagnosis about any abnormality of the subject. All images and reports were acquired from clinical practice, and the reports were written in Chinese. For the sake of conciseness, we name our institutional dataset as TFCH.

Conventionally, despite that each subject undertook examinations of multiple modalities, the ``\textit{Findings}'' section would only focus on describing the specific abnormalities observed from a specific modality. For instance, from the description ``\textit{Skull morphology and size show no abnormalities. In the bilateral basal ganglia, point -like long T1 and long T2 signal shadows are visible. The morphology of the ventricles and sulci, as well as the cisterns, show no abnormal dilation or narrowing, and there is no midline shift. Long T2 signal shadows are visible in the bilateral ethmoid sinuses}'', only T1WI and T2WI are described in the text, while the subject actually undertook examinations including T1WI, T2WI, DWI, ADC, and T2FLAIR. Despite that we can obtain the texts that describe the abnormalities on T1WI and T2WI, we have no knowledge on whether the other modalities, such as DWI, ADC, and T2FLAIR, appear the same as a normal subject. As a result, in this case, we only have image-text pairs for T1WI and T2WI. On the other hand, for subjects without any abnormality, although the texts do not include any modality specific descriptions, we can still assume that all modalities appear normally. As a result, after comparing the images and texts, the number of valid image-text pairs is much smaller than that we originally collected, as summarized in Tab. \ref{tab: pairedimages}. 

After pairing the images and texts, we randomly split the TFCH dataset as a training set with 7825 subjects and a test set with 3000 subjects. We finally obtained 17529 and 3079 image-text pairs in the training and test sets.


\begin{table}[htbp]
    \centering
    \caption{Number of images of each modality in TFCH dataset.}
    \label{tab: modalities}
        \begin{tabular}{cc}
            \hline
            Image modal         &  Number of Images\\ \hline
            T1WI&               10825  \\
            T2WI&               10825  \\
            DWI&                9955 \\
            ADC&                9955 \\
            T2FLAIR&              5952 \\
            \hline
        \end{tabular}
\end{table}

\begin{table}[htbp]
\centering
        \caption{Number of paired images of each modality in TFCH dataset. }
                \label{tab: pairedimages}

        \begin{tabular}{cc}
            \hline
            Image modal         &  Number of Images\\ \hline
            T1WI&               5365  \\
            T2WI&               8424  \\
            DWI&                2838  \\
            ADC&                2839  \\
            T2FLAIR&              1142  \\
            \hline
        \end{tabular}
\end{table}

\subsection{Structural Information Extraction}
\label{sec:Structure Information Extraction}

A radiology report usually includes two sections, i.e., ``\textit{Findings}'' and ``\textit{Impressions}''. 
In ``\textit{Findings}'' section, the descriptions of different modalities are coupled in a single paragraph, sometimes even in a single sentence, making it difficult for image text alignment. 
Moreover, the writing styles vary due to the writing habits among different radiologists. Therefore, it is important to first extract the structural information from the human-written paragraph, so as to better guide the contrastive learning between images and texts.

As the reports were collected from clinical practice, to prevent privacy leakage, we proposed to finetune an LLM for structural information extraction. The LLM takes the original text description as input and the structural information in JSON format as output. 

\begin{figure}[!ht]
\centerline{\includegraphics[width=.4\textwidth]{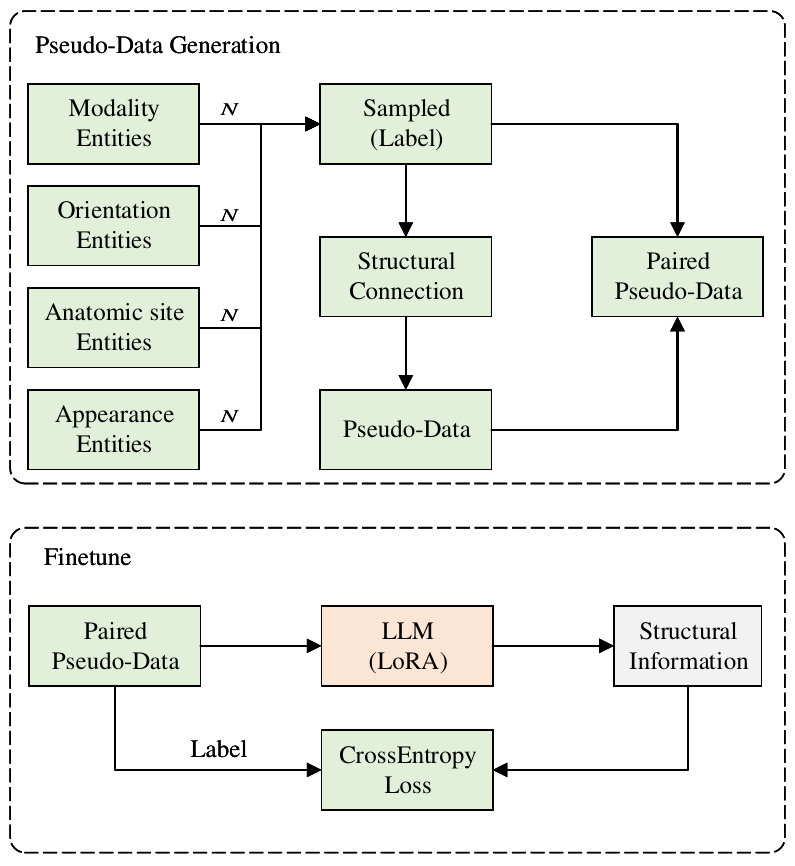}}
\caption{Procedure of structural information extraction using LLM, which includes two phases, i.e., pseudo-data generation and finetune. Pseudo reports are generated from random JSON items, and then used to train the LLM to generate the JSON items from the corresponding pseudo reports.}
\label{fig:finetune LLM}
\end{figure}

The training procedure includes two phases as depicted in Fig. \ref{fig:finetune LLM}, which are pseudo data generation and LLM finetune. In pseudo data generation, we first collected the possible choices of modality, orientation, anatomic site, and appearance. For each pseudo-data sample, we randomly sampled $N$ group of JSON items, where each of which is organized as
\[
\{\text{modality}: m, \text{orientation}: o, \text{anatomic site}: s, \text{appearance}: a \},
\]
where ``modality'' describes from which modality the appearance was found. ``anatomic site'' is where the abnormality was found, such as basal ganglia, temporal lobe gyri, etc.. ``orientation'' describes on which side the abnormality was found, which could be left, right, bilateral, and None.  ``appearance'' describes the exact observations by the radiologists. The group of JSON items is fed into gpt4-turbo, along with an instruction that asks the model to generate the radiology findings according to the given JSON items.

In LLM finetune phase, the generated finding descriptions are used as input, and the corresponding JSON items as output. A Qwen2-7B \cite{yang2024qwen2} pretrained on Chinese texts is adopted as the base model and trained using low-rank adaptation (LoRA).

\subsection{Image-Text Contrastive Learning}
\label{sec:method Image-Text Contrastive Learning}

After extracting the structural information, the descriptions of the abnormalities in each modality was phrased as ``\textit{In modal \{modality\}, at \{orientation\} \{anatomic site\}, the appearance is \{appearance\}.}''
For subjects without any abnormality, a standard expression was used for all modalities, which is ``\textit{The shape and size of the skull are normal. No abnormal signal is observed in the brain parenchyma. The morphology of the ventricles and sulci seen are without abnormal dilation or narrowing, and there is no midline shift. }''

\begin{figure*}[!ht]
\centerline{\includegraphics[width=.7\textwidth]{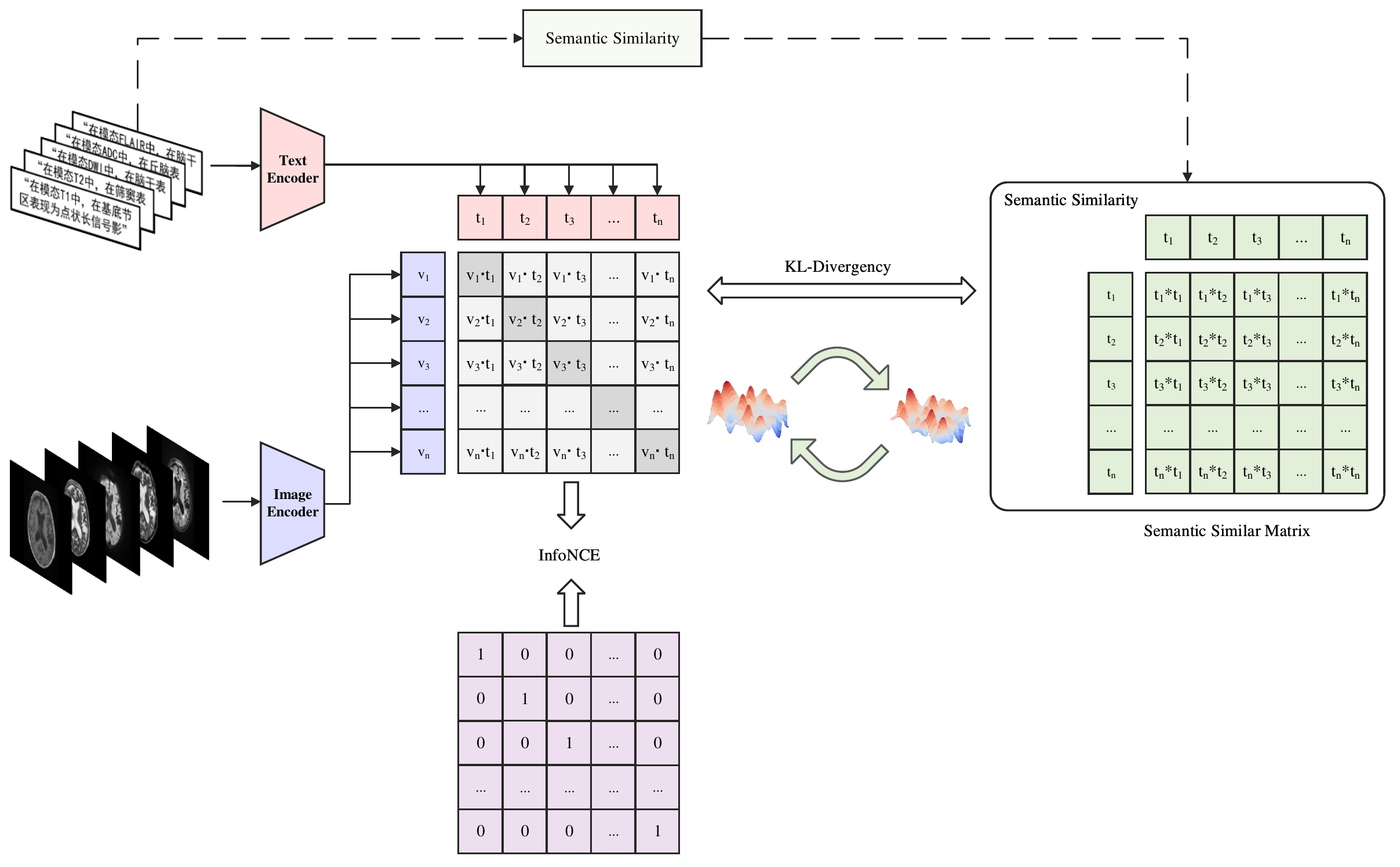}}
\caption{Structure of our proposed SeLIP for multi-modal MRI and radiology findings alignment pretraining.}
\label{fig:train_main}
\end{figure*}

The paired images and texts are then trained by using a CLIP-like training strategy, as described in Fig. \ref{fig:train_main}. The images and the corresponding texts are first encoded as latent vectors, and both image and text encoders are trained to align image and text vectors. 

In particular, denote a batch of image-text pairs as $\{(x_v^i, x_t^i)\}_{i=1}^N$, where $x_v^i$ and $x_t^i$ denote the image and the text of the $i$-th sample in the batch, respectively. The images and the texts are first fed to an image encoder $E_v$ and a text encoder $E_t$, obtain the latent vectors $\{\mathbf{v}_i\}$ and $\{\mathbf{t}_i\}$, respectively. A similarity matrix $\mathbf{C}$ can be then obtained, where $(i,j)$-th element of $\mathbf{C}$ is the cosine similarity between $\mathbf{v}_i$ and $\mathbf{t}_j$, i.e.,
\begin{equation} \label{similarity matrix}
    C_{i,j}=\frac{\mathbf{v}_i\mathbf{t}_j}{\Vert\mathbf{v}_i\Vert\Vert\mathbf{t}_j\Vert}.
\end{equation}
Define image-to-text similarity matrix $\mathbf{P}_{v2t}$ and text-to-image similarity matrix $\mathbf{P}_{t2v}$, their $(i,j)$-th elements can be obtained as
\begin{align}
    p_{v2t}(x_{v}^{i}, x_{t}^{j})=\frac{\exp(C(i, j)/\tau))}{ {\textstyle \sum_{k=1}^{N}(\exp(C(i, k)/\tau))} }, 
\end{align}
and
\begin{equation}
        p_{t2v}(x_{t}^{i}, x_{v}^{j})=\frac{\exp(C(j, i)/\tau))}{ {\textstyle \sum_{k=1}^{N}(\exp(C(k, i)/\tau))} }, 
\end{equation}
respectively, where $\tau$ is the temperature coefficient. 

To train the image and text encoders, the loss function is 
\begin{equation} \label{loss_func}
    L=\alpha L_{clip}+\beta L_{se},
\end{equation}
where $L_{clip}$ is the CLIP loss as described in \cite{radford2021learning}. In particular, we use InfoNCE loss as the CLIP loss. By taking both the image-to-text and text-to-image similarities into consideration, it is defined as
\begin{equation}\label{loss_clip}
    L_{clip}=\frac{L_{v2t}+L_{t2v}}{2},
\end{equation}
where the image-to-text loss $L_{v2t}$ and text-to-image $L_{t2v}$ loss are respectively given as
\begin{equation}
    L_{v2t}= {\textstyle \frac{1}{N} \sum_{i=1}^{N}CE(E_{N},P_{v2t})},
\end{equation}
and
\begin{equation}
    L_{t2v}= {\textstyle \frac{1}{N} \sum_{i=1}^{N}CE(E_{N},P_{t2v})},
\end{equation}
where $CE$ is the cross-entropy loss and $E_{N}$ is an $N\times N$ identity matrix. The second term in (\ref{loss_func}) is the mixed syntax and semantic similarity loss that measures the similarities between texts from both the syntax and semantic level, and will be introduced in detail in the following subsection.

\subsection{Mixed Syntax and Semantic Similarity Matching}
\label{sec:method Semantic Similarity Matching}

\begin{figure}[!ht]
\centerline{\includegraphics[width=0.4\textwidth]{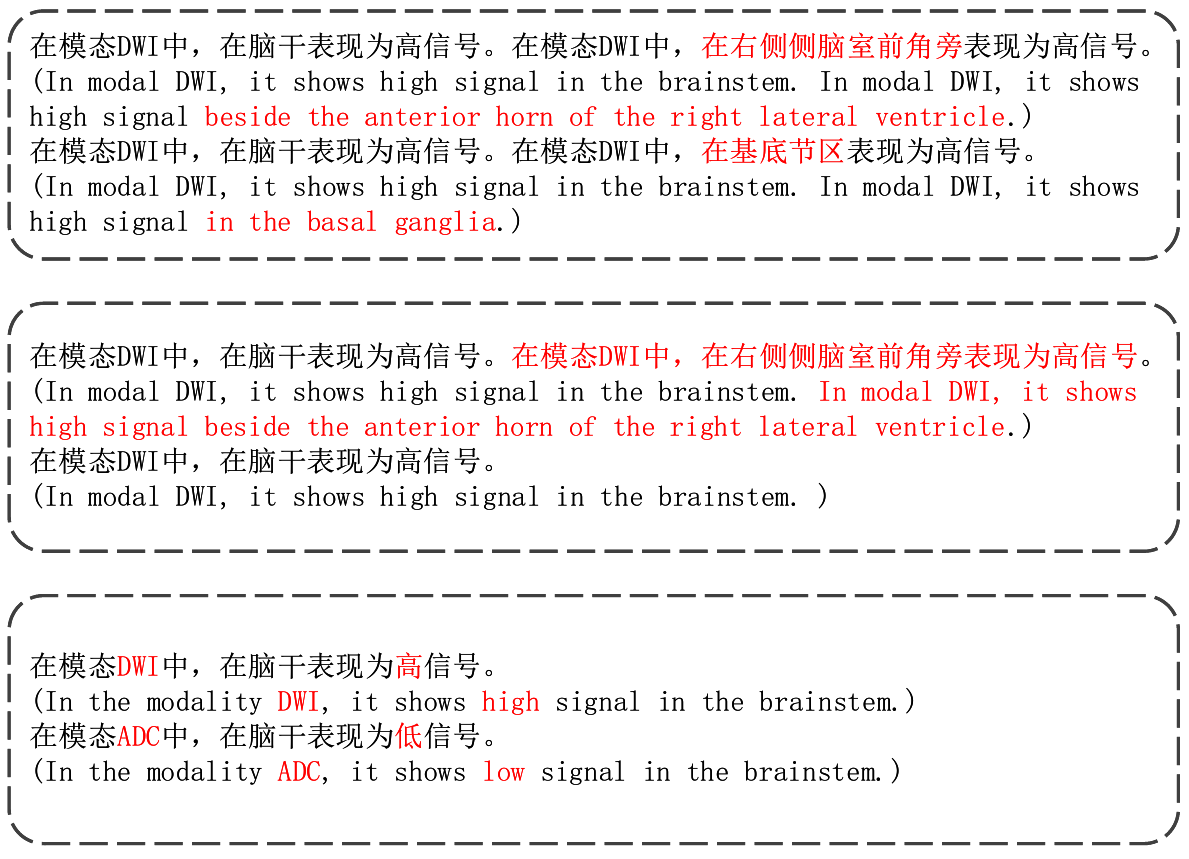}}
\caption{Examples of radiology finding statements pairs that may confuse the network learning.}
\label{fig:chinese_text}
\end{figure}

The text diversity in medical text descriptions is much less than common texts in datasets such as image captioning. Fig.\ref{fig:chinese_text} shows some examples to illustrate similar descriptions on different images. In the first example, both statements include two clauses, where they have one identical clause and one different. In the second example, both statements appear high signal in the brainstem, while the first statement has an additional clause. It is clear that for both cases, it is not appropriate to assign ``0'' as the target in the contrastive learning. In the third example, however, although they have a lot in common in terms of words, they expressed absolutely different semantic meaning. It implies that it is not a good metric to merely compare the differences in words to measure the similarity between two statements.

From the above analysis, we can observe that it is not appropriate to use only the CLIP loss to train the encoders. Instead, it is necessary to introduce an addition metric to measure similarity from both the syntax and the semantic levels between the descriptions of different images. 

The syntax similarity is evaluated using text Dice coefficient (TDC), which is defined as
\begin{equation}
    \text{TDC}=\frac{2  |t_1\bigcap t_2|}{|t_1|+ |t_2|} ,
    \label{eq:TextDice}
\end{equation}
where $t_1$ and $t_2$ denote the clauses to compare. $|t|$ denotes the length of the text. $t_1\bigcap t_2$ denotes the overlapping part of the two clauses. 

For the semantic similarity, we define the site similarity score $w_{loc}$ and appearance similarity score $w_{per}$. $w_{loc}$ measures whether the two clauses are the same in anatomic site, which is assigned to be 1 if they are the same, and 0 otherwise. $w_{per}$ measures whether they describe the same appearance, which is assigned to be 1 if they are the same, and 0 otherwise. Then the mixed syntax and semantic similarity between the two clauses can be obtained as
\begin{equation}
    \tilde{s}_{i,j}=\frac{1}{2}\text{TDC}(t_i, t_j)(w_{loc}(t_i,t_j)+w_{per}(t_i,t_j)).
\end{equation}

Considering that there may be multiple clauses in describing each image, suppose that image $A$ and image $B$ are described by $m$ and $n$ clauses, respectively, the similarity between the descriptions of two images is defined as
\begin{equation}\label{semantic_sim}
    s(A,B)=\frac{1}{mn}\sum_{i=1}^m\sum_{j=1}^n \tilde{s}_{i,j},
\end{equation}
where $\tilde{s}_{i,j}$ denotes the similarity between the $i$-th and the $j$-th clauses in the descriptions of images $A$ and $B$, respectively.

For a batch of data $\{(x_v^i, x_t^i)\}_{i=1}^N$, we compute $s$ according to (\ref{semantic_sim}) between each description pairs, and obtain a semantic similarity matrix $\mathbf{S}$. Kullback–Leibler divergence (KLD) is adopted to encourage the similarity matrix $\mathbf{C}$ defined in (\ref{similarity matrix}) to have the same distribution as $\mathbf{S}$. The image-to-text semantic similarity loss and the text-to-image semantic similarity loss are defined as
\begin{equation}
    L_{se-v2t}= {\textstyle \frac{1}{N} \sum_{i=1}^{N}KL(P_{v2t}\Vert F(\mathbf{S}))},
\end{equation}
and
\begin{equation}
    L_{se-t2v}= {\textstyle \frac{1}{N} \sum_{i=1}^{N}KL(P_{t2v}\Vert F(\mathbf{S}))},
\end{equation}
respectively, where $F(\mathbf{X})$ is the column-wise normalization function to ensure that the sum of elements of each column is 1. The semantic similarity loss can be then obtained as
\begin{equation}
    L_{se}= \frac{1}{2}(L_{se-v2t}+L_{se-t2v}).
\end{equation}

\section{Experiment Results}

\subsection{Implementation}

\subsubsection{Image Preprocessing}

In our TFCH dataset, the image modalities of each subject vary according to clinical need, leading to various image resolutions and intensities. All images were first resampled to a unified matrix size of $24\times 256 \times 256$ by using cubic interpolation, and then truncated the intensity by $99.9$ percentile to remove the outliers. Finally, the images were standardized to $[0,1]$.

\subsubsection{Text and Image Encoders}

In this paper, we adopt BERT \cite{devlin2018bert} as the text encoder, with parameters pretrained on Chinese texts\footnote{https://huggingface.co/google-bert/bert-base-chinese}.A ResNet \cite{he2016deep} with randomly initialized parameters is adopted as the image encoder. The BERT and ResNet outputs, i.e., the image and text embedded vectors, are both projected to the same dimension of 512 before computing the similarities.

\subsubsection{Training Details}

The experiment is conducted on a server with an A800 GPU with Ubuntu ver 20.04. We use PyTorch 2.2.0 and monai 1.3.0 for programming. In our experiment, we set the batch size to be 64, and a sampler is used to ensure that all texts in a single batch are different. We use Adam to optimize both the image and text encoders.  The initial learning rate of image encoder is set to be 0.0001 while the initial learning rate of text encoder is set to be 0.00005. For the image, we adopted extensive data augmentations, including random zooming, random rotation, random contrast adjustment, random scale intensity adjustment, and random elastic deformation. 

We define one epoch as 250 iterations, and the model is trained for 120 epochs. A warming up strategy is adopted, where in the first 5000 iterations (i.e., 20 epochs), at the $t$-th iteration, learning rate is adjusted as
\begin{equation}
     lr^{(t)}=lr_{init}\times \frac{t}{t_{max}},
\end{equation}
where $lr_{init}$ is the initial learning rate, $t_{max}=5000$ is the number of warm-up iterations. After warming up, polynomial learning rate decay is applied after each epoch, and the learning rate of $e$ epoch is calculated as:
\begin{equation}
    lr^{(e)}=lr_{init}\times (1-\frac{e}{e_{max}})^{0.9} ,
\end{equation}
where $e_{max}$ is maximum number of epochs after warming-up stage, and is set to 100 in our experiments.

\subsection{Performance of Structural Information Extraction}

We first analyze the effectiveness of structural information extraction by LLMs. We use JSON parsing success rate to evaluate whether the LLM can generate legal JSON format. We also use 200 real clinical reports to evaluate the accuracy, where a generated JSON item is said to be correct if it is correct in modality, orientation, anatomic site, and appearance.The results are summarized in Tab.\ref{tab: Parsing success rate of LLMs}. For the sake of comparison, the results generated by a gpt4-turbo, as well as a finetuned ChatGLM-6B\cite{du2021glm}, are also presented. The parsing success rate is defined as ratio of the number of successfully parsed radiology reports to the number of total reports. The accuracy is defined at JSON item level, which is defined as
\begin{equation}
    acc=\frac{TP}{TP+FP+FN},
\end{equation}
where $TP$ is the correctly generated number of items, $FN$ is the number of missing items, and $FP$ is the number of items that generated by the LLM but actually do not exist in the reports.

As we can see from Tab.\ref{tab: Parsing success rate of LLMs}, gpt4-turbo presented the worst JSON parsing success rate due to the complicated and diverse writting styles in radiology reports. The locally finetuned LLMs, i.e., ChatGLM and Qwen2, presented very high parsing success rate. On the other hand, the accuracies of the locally finetuned LLMs are also much higher than gpt4-turbo. Qwen2-7B presented the best performance, which achieved an accuracy of $94.59\%$. We can then conclude that Qwen2-7B can be used in structural information extraction from human-written reports with high efficiency and high accuracy.

\begin{table}[!ht]
    \centering
    \caption{Parsing success rate and accuracy of LLMs. }
    \label{tab: Parsing success rate of LLMs}
    \begin{tabular}{c|cc}
    \hline
    Model  & Parsing success rate  & Accuracy  \\ \hline

    GPT-4-turbo    & 0.7700                & 0.5238  \\
    ChatGLM-6B-finetuned    & 0.9900       & 0.8604  \\
    Qwen2-7B-finetuned  & 0.9950           & 0.9459  \\
    \hline

    \end{tabular}
\end{table}

\subsection{Image Text Retrieval}
\label{sec:experiment--Image Text Retrieval}

\begin{figure*}[!ht]
\centerline{\includegraphics[width=.8\textwidth]{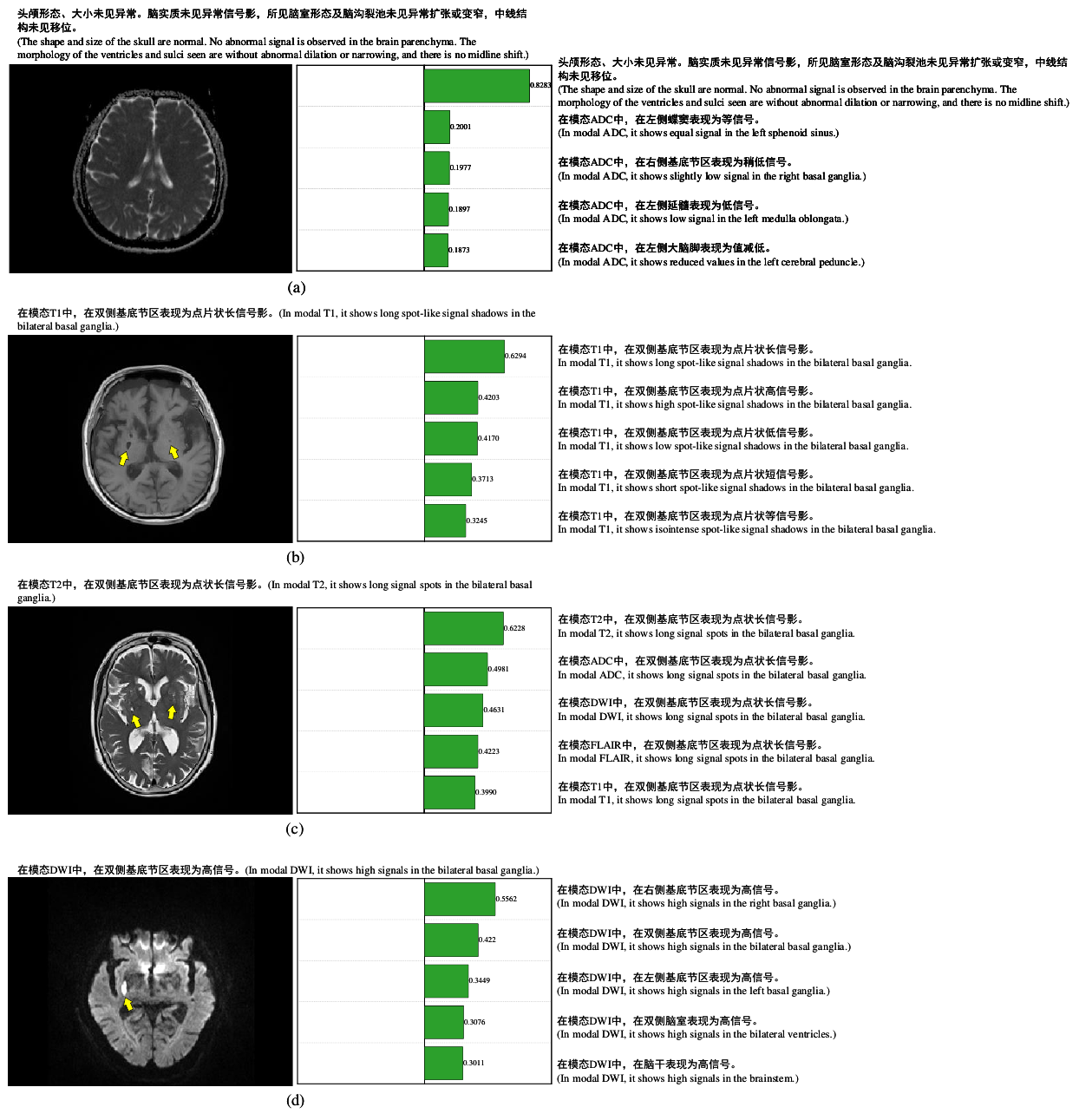}}
\caption{Examples of the cosine similarity measurements between an image and candidate text descriptions by using ResNet50 as backbone. The real lesions on the image are identified by yellow arrows.}
\label{fig:image_find_text}
\end{figure*}

The performance of image text alignment is evaluated by testing the image-text retrieval on our test set. We compute cosine similarities between the input image and all candidate texts, and then sort the texts according to the similarities. Fig.\ref{fig:image_find_text} presents some examples of the image-text retrieval. In the first example, we can see that our pretrained model presents an excellent ability to match the image of the healthy sample and the healthy text. The second example shows that our pretrained model to distinguish the appearance of the image. The third example shows that our pretrained model can distinguish among all candidate modalities. The last example shows its ability to distinguish the orientation and anatomic site. 

To better illustrate the image-text retrieval ability, we evaluate the performance on our test set, as summarized in Tab. \ref{tab: image-to-text retrieval}, where Top-K accuracy means that the correct text is among the texts within the $K$-th highest similarity scores. Note that although there are 3079 image-text pairs in the test set, we only have 1689 non-duplicated texts. Therefore, the embedding vector of each image was compared with all 1689 non-duplicated texts. As we can see from Tab. \ref{tab: image-to-text retrieval}, our proposed SeLIP presents better performance than the conventional CLIP model in the Top-1 to Top-10 accuracies.

\begin{table}[!ht]
    \centering
    \caption{Performance of image-to-text retrieval task on TFCH test set. }
    \label{tab: image-to-text retrieval}
    \begin{tabular}{c|ccccc}
    \hline
    Backbone  & Method  & Top-1 & Top-2 & Top-5  & Top-10  \\ \hline

    \multirow{2}{*}{ResNet18}  & CLIP  & 0.2059  & 0.2342  & 0.2686  & 0.3007 \\ 
    & SeLIP  & 0.2809  & 0.2897  & 0.3134  & 0.3423 \\ \hline
    
    \multirow{2}{*}{ResNet50}  & CLIP  & 0.2283	& 0.2514	& 0.2887	& 0.3303 \\
    & SeLIP  & 0.2871	 & 0.2985  & 0.3212  & 0.3514 \\ \hline

    \end{tabular}
\end{table}

\subsection{Image Encoder as a Classifier's Backbone}\label{sec_cls}

To evaluate the performance of our pretraining method, we evaluate the performance of the image encoder as the backbone network of a classifier. The classifier is constructed by adding a fully-connected (FC) layer after the global average pooling layer of the ResNet. In this subsection, ResNet18 and ResNet50 are adopted as backbones, and three experiment settings are considered as follows.
\begin{itemize}
    \item Zero-Shot (ZS). The parameters of the pretrained ResNet backbone are frozen, and only the newly added FC layer is trained.
    \item Finetuning (FT). The pretrained parameters are loaded. All layers are trained.
    \item Train-From-Scratch (TFS). All parameters are randomly initiated, and all layers are trained.
\end{itemize}

\subsubsection{Classifying Data From the Same Domain}
\label{sec:experiment--Image Classification in Same Domain}

We collected another set of head DWI images from the same institute, i.e., Tianjin First Center Hospital. In particular, we collected 1200 DWI images, where half of them are diagnosed as acute ischemic stroke. The images are randomly divided as a training set of 800 samples, and a test set of 400 samples. 

Tab. \ref{tab: stroke results} summarizes the experiment results. For the sake of comparison, the results of a CLIP model are also presented, where only $L_{clip}$ defined in (\ref{loss_clip}) is used to train the model. Moreover, to evaluate the performance with a small training set, the performances where the classifiers are only trained on 10\% and 25\% training samples are also provided. 

We can see from Tab. \ref{tab: stroke results} that in both ResNet18 and ResNet50 cases, almost all ZS models and FT models achieve better performance than their corresponding TFS models. When the number of training samples is small, the models with pretrained parameters present much better performance, which proves that the pretrained image encoder is able to extract features from the contrastive learning. More importantly, our proposed SeLIP outperforms CLIP in almost all cases, which highlights the effectiveness of our proposed mixed syntax and semantic similarity loss.

\subsubsection{Classifying Cross Domain Data}
\label{sec:Classifying Cross Domain Data}

To evaluate the generalization of our pretrained image encoder, we further trained a classifier on cross domain data, i.e., BraTS19 dataset \cite{bakas2018identifying}. In particular, BraTS19 dataset includes 332 subjects with four modalities, i.e., T1WI, T2WI, FLAIR, and contrast-enhanced T1WI (T1ce). In additional to the tumor segmentation masks, the dataset also provide labels about whether the subject's tumor is low-grade glioma (LGG) or high-grade glioma (HGG). The dataset includes 259 HGG subjects and 73 LGG ones. 
Following \cite{chatterjee2022classification}, we split the training and test datasets as 7:3. T1ce, a modality that has never been seen by our pretrained image encoder, is used to train a classifier that distinguishes LGG from HGG. 

\begin{table*}[!ht]
    \centering
    \caption{Results of stroke classification on the data from the same data domain. }
    \label{tab: stroke results}
    \begin{tabular}{c|ccc|ccc|ccc}
    \hline
    \multirow{2}{*}{Method}    & \multicolumn{3}{c}{10\% Training Set}    & \multicolumn{3}{c}{25\% Training Set}    & \multicolumn{3}{c}{100\% Training Set} \\
    \cline{2-10}
    & ACC  & F1  & AUC  & ACC  & F1  & AUC  & ACC  & F1  & AUC  \\
    \hline
    ResNet18-TFS          & 82.00  & 82.18  & 89.51  & 85.50  & 85.43  & 92.74  & 86.75  & 87.04  & 93.49 \\
    ResNet18-CLIP-ZS      & 84.50  & 83.42  & 90.27  & 86.25  & 85.64  & 90.71  & 85.50  & 84.07  & 91.75  \\
    ResNet18-CLIP-FT      & 87.25  & 86.75  & 91.33  & 88.75  & 88.06  & 92.54  & 89.00  & 88.89  & 93.43  \\
    ResNet18-SeLIP-ZS  & 87.75  & 87.96  & 93.18  & 87.50  & 87.44  & 93.60  & 87.75  & 87.34  & 93.81  \\
    ResNet18-SeLIP-FT  & 86.75  & 86.78  & 91.48  & 88.25  & 87.92  & 93.18  & 91.00  & 90.53  & 95.34  \\

    \hline
    ResNet50-TFS          & 79.75  & 78.85  & 85.96  & 83.25  & 82.23  & 91.91  & 86.75  & 86.58  & 91.65 \\
    ResNet50-CLIP-ZS      & 83.00  & 83.57  & 89.11  & 83.50  & 82.54  & 89.48  & 84.25  & 83.89  & 90.37  \\
    ResNet50-CLIP-FT      & 85.75  & 85.42  & 91.41  & 87.25  & 87.22  & 92.31  & 88.75  & 88.89  & 93.60  \\
    ResNet50-SeLIP-ZS  & 85.50  & 84.82  & 91.20  & 86.00  & 84.95  & 91.85  & 87.25  & 86.54  & 92.34  \\
    ResNet50-SeLIP-FT  & 85.75  & 85.35  & 90.48  & 88.25  & 87.73  & 93.07  & 90.00  & 89.95  & 93.49  \\

    \hline

    \end{tabular}
\end{table*}

\begin{table*}[!ht]
    \centering
    \caption{Results of classifying HGG from LGG on the T1ce images of BraTS19 dataset.}
    \label{tab: brats zs only t1ce}
    \begin{tabular}{c|ccc|ccc|ccc}
    \hline
    \multirow{2}{*}{Method}    & \multicolumn{3}{c}{10\% Training Set}    & \multicolumn{3}{c}{40\% Training Set}    & \multicolumn{3}{c}{100\% Training Set} \\
    \cline{2-10}
    & ACC  & F1  & AUC  & ACC  & F1  & AUC  & ACC  & F1  & AUC  \\
    \hline
       
    DSM\cite{chatterjee2022classification}  &-  &-  &-  &-  &-  &-  &-   &90.36    &-  \\
    KAD-ZS\cite{zhang2023knowledge} &-  &-  &-  &-  &-  &-    & 51.48    & 62.54    & 50.52   \\
    UniBrain-ZS\cite{lei2023unibrain} &-  &-  &-  &-  &-  &-  & 59.11  & 64.88  & 62.77  \\
       
    \hline
    ResNet18-TFS          & 85.15  & 90.20  & 86.79  & 87.13  & 91.93  & 87.90  & 86.14  & 91.03  & 88.96 \\
    ResNet18-CLIP-ZS      & 71.29  & 78.20  & 82.83  & 73.27  & 80.00  & 82.66  & 75.25  & 82.01  & 86.45 \\
    ResNet18-SeLIP-ZS  & 71.29  & 79.14  & 73.08  & 74.26  & 82.19  & 76.64  & 76.24  & 84.21  & 74.75 \\
    ResNet18-CLIP-FT      & 81.19  & 87.90 & 83.67  & 87.13  & 91.61  & 90.25  & 85.15  & 91.12  & 87.18 \\
    ResNet18-SeLIP-FT  & 85.15  & 90.45  & 84.34  & 86.14  & 91.25  & 90.30  & 89.11  & 93.33  & 91.19 \\

    \hline
    ResNet50-TFS          & 85.15  & 90.45  & 79.15  & 87.13  & 91.50  & 87.40  & 88.12  & 92.31  & 86.96 \\
    ResNet50-CLIP-ZS      & 80.20  & 86.67  & 82.27  & 77.23  & 84.35  & 81.88  & 81.19  & 86.71  & 84.34 \\
    ResNet50-SeLIP-ZS  & 84.16  & 89.61  & 85.79  & 85.15  & 90.45  & 85.51  & 85.15  & 90.32  & 83.50 \\
    ResNet50-CLIP-FT      & 83.17  & 89.70  & 79.21  & 87.13  & 91.93  & 86.29  & 85.15  & 90.20  & 87.85 \\
    ResNet50-SeLIP-FT  & 86.14  & 91.46  & 84.23  & 89.11  & 93.08  & 88.71  & 90.10  & 93.59  & 92.92 \\

    \hline

    \end{tabular}
\end{table*}

Tab. \ref{tab: brats zs only t1ce} presents the evaluation results on the test set. As we can see, our proposed SeLIP is able to achieve high accuracy even when the number of training samples is small. Comparing to the DSM which is a TFS model, SeLIP-ZS model achieved close performance, while SeLIP-FT has a improvement in F1 by 3.23\%. When compared to KAD, there are 35.65\%, 27.78\% and 32.98\% improvement of SeLIP-ZS model in ACC, F1 and AUC, respectively. Our experiment on BraTS19 suggests that our proposed SeLIP framework has a strong cross-domain ability even on a modality that has never been seen before.

\subsection{Image Encoder as a Segmentor’s Encoder}

We further build a segmentation network based on the pretrained image encoder. The global average pooling layer of the ResNet is removed, and an FCN-8s head \cite{long2015fully} is added. Similar to Sec. \ref{sec_cls}, three experiments are conducted as follows. 
\begin{itemize}
    \item ZS. The parameters of the pretrained ResNet backbone are frozen, and only the newly added FCN8s decoder is trained.
    \item FT. The pretrained parameters are loaded. All layers are trained.
    \item TFS. All parameters are randomly initiated, and all layers are trained.
\end{itemize}

\begin{figure*}[!ht]
\centerline{\includegraphics[width=\textwidth]{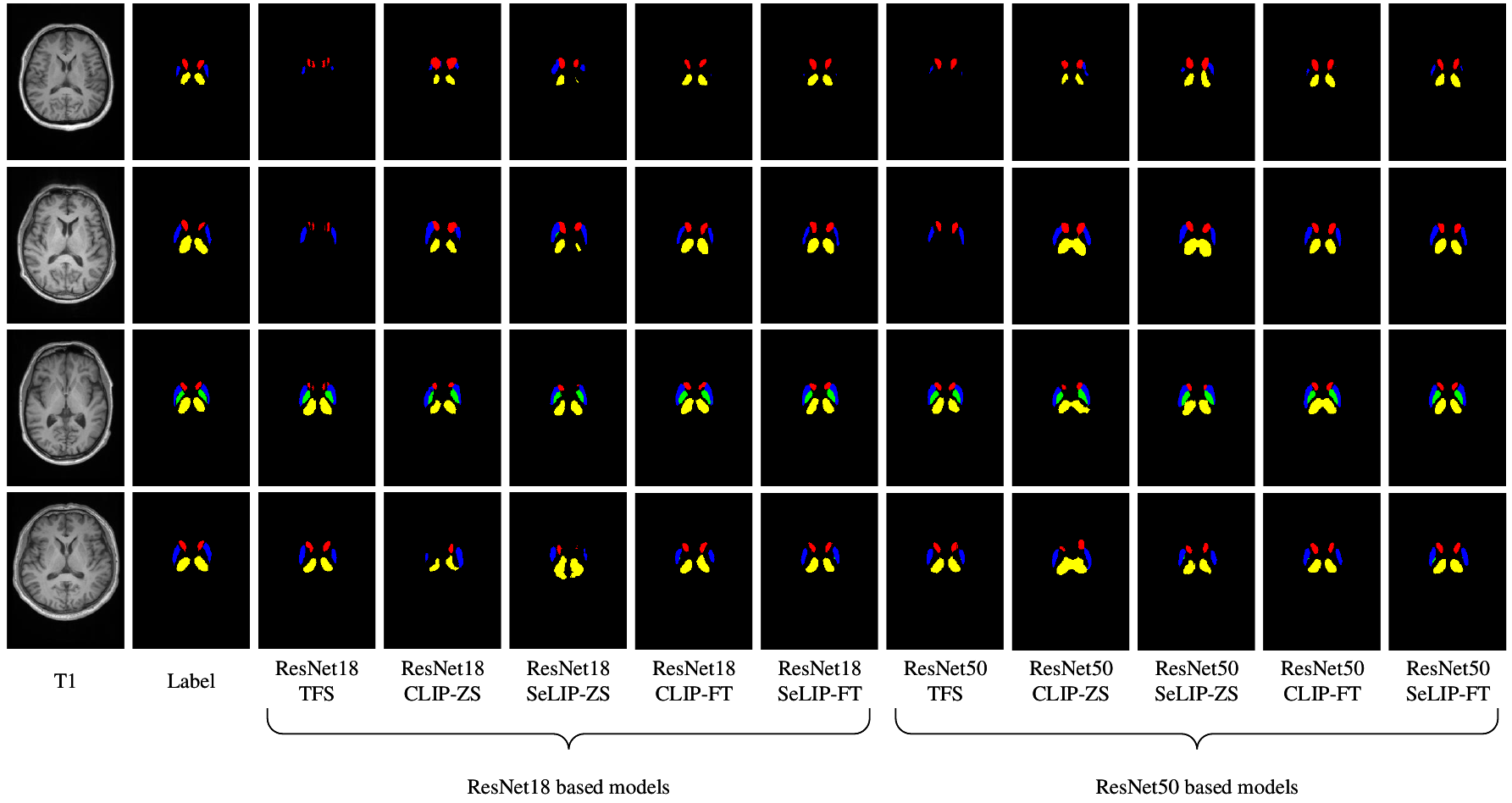}}
\caption{Examples of brain nuclei segmentation. The segmentation masks are rendered in color, where the colors of CN, GP, PUT, THA are shown as red, green, blue, yellow, respectively. The first and second columns are the original T1WI and nuclei labels. Columns 3 to 7 show the segmentation results with ResNet18 backbone under various training strategies. Columns 8 to 12 show the results with ResNet50 backbone.}
\label{fig:seg_results}
\end{figure*}

We collect 27 head T1 images from Tianjin First Central Hospital, and the dataset is divided as a training set with 20 samples and a test set with 7 samples. Four different gray matter nuclei regions are labeled in images, including the bilateral caudate nuclei (CN), globus pallidus (GP), putamen (PUT), and thalamus (THA). All images are first resampled from $64\times 288 \times 384$ to $24\times 256 \times 256$. The preprocessing methods are the same as pretraining data, including intensity truncating by $99.9$ percentile and standardizing to $[0,1]$. Adam optimizer was used during training, where the initial learning rate is set to 0.0001 and a polynomial learning rate decay is applied. 

\begin{table}[!ht]
    \centering
    \caption{Dice coefficient of segmentation on the test set.}
    \label{tab: segmentation}
    \begin{tabular}{c|ccccc}
    \hline
    Method  & CN  & GP  & PUT  & THA  & Average \\
    \hline
     ResNet18-TFS  & 0.6120  &0.7674  & 0.7208  & 0.7128  & 0.7033 \\   
       ResNet18-CLIP-ZS  & 0.5456  & 0.6757  & 0.6069  & 0.7259  & 0.6385 \\   
       ResNet18-SeLIP-ZS & 0.5315  & 0.6759  & 0.6170  & 0.6683  & 0.6232 \\   
       ResNet18-CLIP-FT  & 0.7174  & 0.7406  & 0.6833  & 0.8161  & 0.7393 \\   
       ResNet18-SeLIP-FT  & 0.7348  & 0.7794  & 0.6975  & 0.8207  & 0.7581 \\   
    \hline
       ResNet50-TFS  & 0.7499  & 0.6876  & 0.7122  & 0.7282  & 0.7195 \\   
       ResNet50-CLIP-ZS  & 0.6098  & 0.7375  & 0.6727  & 0.7731  & 0.6983 \\   
       ResNet50-SeLIP-ZS & 0.6700  & 0.7196  & 0.6871  & 0.7499  & 0.7066 \\   
       ResNet50-CLIP-FT  & 0.7228  & 0.7960  & 0.6929  & 0.7984  & 0.7525 \\   
       ResNet50-SeLIP-FT  & 0.7300  & 0.7715  & 0.7190  & 0.8172  & 0.7594 \\ 
    \hline

    \end{tabular}
\end{table}

Fig. \ref{fig:seg_results} presents some examples from the test set. As we can see from Fig. \ref{fig:seg_results}, the proposed contrastive learning method is able to segment the nuclei even when only the decoder head is trained. The performance can be further improved when all layers of the encoder and the decoder are trained. Tab. \ref{tab: segmentation} further summarized the Dice coefficients on the test set. As we can see from Tab. \ref{tab: segmentation}, ZS is able to achieve close performance to TFS when using ResNet50 backbone. When all parameters are trained from the pretrained encoder, FT is 5.48\% and 3.99\% better than the models trained from randomly initiated parameters (TFS). The performance indicates that although the model is trained with the target of aligning images and the corresponding text descriptions, the image encoder is able to preserve adequate spatial information and can be easily transferred to segmentation tasks.

\section{Conclusion}

In this paper, we proposed a contrastive learning method for image encoder pretraining on multi-modal head MRI and Chinese radiology reports. By integrating the mixed syntax and semantic similarity metric to the conventional CLIP framework, our proposed SeLIP is able to effectively extract more useful features. Experiments revealed that our proposed SeLIP performs well in many downstream tasks including image-text retrieval task, classification tasks, and segmentation task, which highlights the importance of enhancing the contrastive learning by considering the similarities among texts describing different images. Our proposed SeLIP shed a light on training foundation models for medical images.

While the transformers, such as Vision Transformer (ViT), are commonly believed to be more suitable for image-text alignment, it should be noted that compared with the works on common images, the number of paired image and texts is relatively small. We are working on collecting more images and texts, so as to develop a more generalized pretrained model in our future work.

\section*{Acknowledgement}
This work is supported in part by Tianjin Municipal Education Committee (grant No. B231005531), Tianjin Health High Level Talent Selection and Training Project (TJSQNYXXR-D2-143), Tianjin Health Research Project (TJWJ2023QN031), and Tianjin Key Medical Discipline (Specialty) Construction Project (TJYXZDXK-041A).

\printcredits

\bibliographystyle{unsrtnat}

\bibliography{references}

\begin{thebibliography}{52}
\providecommand{\natexlab}[1]{#1}
\providecommand{\url}[1]{\texttt{#1}}
\expandafter\ifx\csname urlstyle\endcsname\relax
  \providecommand{\doi}[1]{doi: #1}\else
  \providecommand{\doi}{doi: \begingroup \urlstyle{rm}\Url}\fi

\bibitem[Li et~al.(2023)Li, Li, Yan, Li, Jiang, Luo, and Yin]{li2023deep}
Xiang Li, Minglei Li, Pengfei Yan, Guanyi Li, Yuchen Jiang, Hao Luo, and Shen Yin.
\newblock Deep learning attention mechanism in medical image analysis: Basics and beyonds.
\newblock \emph{International Journal of Network Dynamics and Intelligence}, pages 93--116, 2023.

\bibitem[Rana and Bhushan(2023)]{rana2023machine}
Meghavi Rana and Megha Bhushan.
\newblock Machine learning and deep learning approach for medical image analysis: diagnosis to detection.
\newblock \emph{Multimedia Tools and Applications}, 82\penalty0 (17):\penalty0 26731--26769, 2023.

\bibitem[Mall et~al.(2023)Mall, Singh, Srivastav, Narayan, Paprzycki, Jaworska, and Ganzha]{mall2023comprehensive}
Pawan~Kumar Mall, Pradeep~Kumar Singh, Swapnita Srivastav, Vipul Narayan, Marcin Paprzycki, Tatiana Jaworska, and Maria Ganzha.
\newblock A comprehensive review of deep neural networks for medical image processing: Recent developments and future opportunities.
\newblock \emph{Healthcare Analytics}, page 100216, 2023.

\bibitem[Ronneberger et~al.(2015)Ronneberger, Fischer, and Brox]{ronneberger2015u}
Olaf Ronneberger, Philipp Fischer, and Thomas Brox.
\newblock U-net: Convolutional networks for biomedical image segmentation.
\newblock In \emph{Medical image computing and computer-assisted intervention--MICCAI 2015: 18th international conference, Munich, Germany, October 5-9, 2015, proceedings, part III 18}, pages 234--241. Springer, 2015.

\bibitem[Hatamizadeh et~al.(2022)Hatamizadeh, Tang, Nath, Yang, Myronenko, Landman, Roth, and Xu]{hatamizadeh2022unetr}
Ali Hatamizadeh, Yucheng Tang, Vishwesh Nath, Dong Yang, Andriy Myronenko, Bennett Landman, Holger~R Roth, and Daguang Xu.
\newblock Unetr: Transformers for 3d medical image segmentation.
\newblock In \emph{Proceedings of the IEEE/CVF winter conference on applications of computer vision}, pages 574--584, 2022.

\bibitem[Ali et~al.(2022)Ali, Ahmed, Paul, Jahan, Sani, Noor, and Hasan]{ali2022monkeypox}
Shams~Nafisa Ali, Md~Tazuddin Ahmed, Joydip Paul, Tasnim Jahan, SM~Sani, Nawsabah Noor, and Taufiq Hasan.
\newblock Monkeypox skin lesion detection using deep learning models: A feasibility study.
\newblock \emph{arXiv preprint arXiv:2207.03342}, 2022.

\bibitem[Jiang et~al.(2023)Jiang, Hu, Wang, and Zhang]{jiang2023deep}
Xiaoyan Jiang, Zuojin Hu, Shuihua Wang, and Yudong Zhang.
\newblock Deep learning for medical image-based cancer diagnosis.
\newblock \emph{Cancers}, 15\penalty0 (14):\penalty0 3608, 2023.

\bibitem[Mijwil et~al.(2023)Mijwil, Al-Mistarehi, Abotaleb, El-kenawy, Ibrahim, Abdelhamid, and Eid]{mijwil2023pixels}
Maad~M Mijwil, Abdel-Hameed Al-Mistarehi, Mostafa Abotaleb, El-Sayed~M El-kenawy, Abdelhameed Ibrahim, Abdelaziz~A Abdelhamid, and Marwa~M Eid.
\newblock From pixels to diagnoses: Deep learning's impact on medical image processing-a survey.
\newblock \emph{Wasit Journal of Computer and Mathematics Science}, 2\penalty0 (3):\penalty0 9--15, 2023.

\bibitem[Wang et~al.(2020)Wang, Zhang, Lam, Cai, and Yang]{wang2020review}
Mingqing Wang, Qilin Zhang, Saikit Lam, Jing Cai, and Ruijie Yang.
\newblock A review on application of deep learning algorithms in external beam radiotherapy automated treatment planning.
\newblock \emph{Frontiers in oncology}, 10:\penalty0 580919, 2020.

\bibitem[Shen et~al.(2020)Shen, Nguyen, Zhou, Jiang, Dong, and Jia]{shen2020introduction}
Chenyang Shen, Dan Nguyen, Zhiguo Zhou, Steve~B Jiang, Bin Dong, and Xun Jia.
\newblock An introduction to deep learning in medical physics: advantages, potential, and challenges.
\newblock \emph{Physics in Medicine \& Biology}, 65\penalty0 (5):\penalty0 05TR01, 2020.

\bibitem[Deng et~al.(2009)Deng, Dong, Socher, Li, Li, and Fei-Fei]{deng2009imagenet}
Jia Deng, Wei Dong, Richard Socher, Li-Jia Li, Kai Li, and Li~Fei-Fei.
\newblock Imagenet: A large-scale hierarchical image database.
\newblock In \emph{2009 IEEE conference on computer vision and pattern recognition}, pages 248--255. Ieee, 2009.

\bibitem[He et~al.(2022)He, Chen, Xie, Li, Doll{\'a}r, and Girshick]{he2022masked}
Kaiming He, Xinlei Chen, Saining Xie, Yanghao Li, Piotr Doll{\'a}r, and Ross Girshick.
\newblock Masked autoencoders are scalable vision learners.
\newblock In \emph{Proceedings of the IEEE/CVF conference on computer vision and pattern recognition}, pages 16000--16009, 2022.

\bibitem[Liu et~al.(2021)Liu, Zhang, Hou, Mian, Wang, Zhang, and Tang]{liu2021self}
Xiao Liu, Fanjin Zhang, Zhenyu Hou, Li~Mian, Zhaoyu Wang, Jing Zhang, and Jie Tang.
\newblock Self-supervised learning: Generative or contrastive.
\newblock \emph{IEEE transactions on knowledge and data engineering}, 35\penalty0 (1):\penalty0 857--876, 2021.

\bibitem[Jaiswal et~al.(2020)Jaiswal, Babu, Zadeh, Banerjee, and Makedon]{jaiswal2020survey}
Ashish Jaiswal, Ashwin~Ramesh Babu, Mohammad~Zaki Zadeh, Debapriya Banerjee, and Fillia Makedon.
\newblock A survey on contrastive self-supervised learning.
\newblock \emph{Technologies}, 9\penalty0 (1):\penalty0 2, 2020.

\bibitem[Chen et~al.(2020)Chen, Li, Yu, El~Kholy, Ahmed, Gan, Cheng, and Liu]{chen2020uniter}
Yen-Chun Chen, Linjie Li, Licheng Yu, Ahmed El~Kholy, Faisal Ahmed, Zhe Gan, Yu~Cheng, and Jingjing Liu.
\newblock Uniter: Universal image-text representation learning.
\newblock In \emph{European conference on computer vision}, pages 104--120. Springer, 2020.

\bibitem[Devlin(2018)]{devlin2018bert}
Jacob Devlin.
\newblock Bert: Pre-training of deep bidirectional transformers for language understanding.
\newblock \emph{arXiv preprint arXiv:1810.04805}, 2018.

\bibitem[Radford et~al.(2021)Radford, Kim, Hallacy, Ramesh, Goh, Agarwal, Sastry, Askell, Mishkin, Clark, et~al.]{radford2021learning}
Alec Radford, Jong~Wook Kim, Chris Hallacy, Aditya Ramesh, Gabriel Goh, Sandhini Agarwal, Girish Sastry, Amanda Askell, Pamela Mishkin, Jack Clark, et~al.
\newblock Learning transferable visual models from natural language supervision.
\newblock In \emph{International conference on machine learning}, pages 8748--8763. PMLR, 2021.

\bibitem[Harnoune et~al.(2021)Harnoune, Rhanoui, Mikram, Yousfi, Elkaimbillah, and El~Asri]{harnoune2021bert}
Ayoub Harnoune, Maryem Rhanoui, Mounia Mikram, Siham Yousfi, Zineb Elkaimbillah, and Bouchra El~Asri.
\newblock Bert based clinical knowledge extraction for biomedical knowledge graph construction and analysis.
\newblock \emph{Computer Methods and Programs in Biomedicine Update}, 1:\penalty0 100042, 2021.

\bibitem[Meoni et~al.(2023)Meoni, Ryffel, and de~la Clergerie]{meoni2023large}
Simon Meoni, Theo Ryffel, and Eric de~la Clergerie.
\newblock Large language models as instructors: A study on multilingual clinical entity extraction.
\newblock In \emph{The 22nd Workshop on Biomedical Natural Language Processing and BioNLP Shared Tasks}, pages 178--190. Association for Computational Linguistics, 2023.

\bibitem[Wang et~al.(2022)Wang, Wu, Agarwal, and Sun]{wang2022medclip}
Zifeng Wang, Zhenbang Wu, Dinesh Agarwal, and Jimeng Sun.
\newblock Medclip: Contrastive learning from unpaired medical images and text.
\newblock \emph{arXiv preprint arXiv:2210.10163}, 2022.

\bibitem[Gao et~al.(2024)Gao, Liu, Xu, Wu, Zhang, Li, Yang, Liu, and Sun]{gao2024softclip}
Yuting Gao, Jinfeng Liu, Zihan Xu, Tong Wu, Enwei Zhang, Ke~Li, Jie Yang, Wei Liu, and Xing Sun.
\newblock Softclip: Softer cross-modal alignment makes clip stronger.
\newblock In \emph{Proceedings of the AAAI Conference on Artificial Intelligence}, volume~38, pages 1860--1868, 2024.

\bibitem[Jia et~al.(2021)Jia, Yang, Xia, Chen, Parekh, Pham, Le, Sung, Li, and Duerig]{jia2021scaling}
Chao Jia, Yinfei Yang, Ye~Xia, Yi-Ting Chen, Zarana Parekh, Hieu Pham, Quoc Le, Yun-Hsuan Sung, Zhen Li, and Tom Duerig.
\newblock Scaling up visual and vision-language representation learning with noisy text supervision.
\newblock In \emph{International conference on machine learning}, pages 4904--4916. PMLR, 2021.

\bibitem[Pham et~al.(2023)Pham, Dai, Ghiasi, Kawaguchi, Liu, Yu, Yu, Chen, Luong, Wu, et~al.]{pham2023combined}
Hieu Pham, Zihang Dai, Golnaz Ghiasi, Kenji Kawaguchi, Hanxiao Liu, Adams~Wei Yu, Jiahui Yu, Yi-Ting Chen, Minh-Thang Luong, Yonghui Wu, et~al.
\newblock Combined scaling for zero-shot transfer learning.
\newblock \emph{Neurocomputing}, 555:\penalty0 126658, 2023.

\bibitem[Zhai et~al.(2022)Zhai, Wang, Mustafa, Steiner, Keysers, Kolesnikov, and Beyer]{zhai2022lit}
Xiaohua Zhai, Xiao Wang, Basil Mustafa, Andreas Steiner, Daniel Keysers, Alexander Kolesnikov, and Lucas Beyer.
\newblock Lit: Zero-shot transfer with locked-image text tuning.
\newblock In \emph{Proceedings of the IEEE/CVF conference on computer vision and pattern recognition}, pages 18123--18133, 2022.

\bibitem[Cherti et~al.(2023)Cherti, Beaumont, Wightman, Wortsman, Ilharco, Gordon, Schuhmann, Schmidt, and Jitsev]{cherti2023reproducible}
Mehdi Cherti, Romain Beaumont, Ross Wightman, Mitchell Wortsman, Gabriel Ilharco, Cade Gordon, Christoph Schuhmann, Ludwig Schmidt, and Jenia Jitsev.
\newblock Reproducible scaling laws for contrastive language-image learning.
\newblock In \emph{Proceedings of the IEEE/CVF Conference on Computer Vision and Pattern Recognition}, pages 2818--2829, 2023.

\bibitem[Li et~al.(2021)Li, Liang, Zhao, Cui, Ouyang, Shao, Yu, and Yan]{li2021supervision}
Yangguang Li, Feng Liang, Lichen Zhao, Yufeng Cui, Wanli Ouyang, Jing Shao, Fengwei Yu, and Junjie Yan.
\newblock Supervision exists everywhere: A data efficient contrastive language-image pre-training paradigm.
\newblock \emph{arXiv preprint arXiv:2110.05208}, 2021.

\bibitem[Shi et~al.(2025)Shi, Zhao, Wang, Zhang, Wang, Li, Dai, Zou, Xiong, Tian, et~al.]{shi2025umg}
Bowen Shi, Peisen Zhao, Zichen Wang, Yuhang Zhang, Yaoming Wang, Jin Li, Wenrui Dai, Junni Zou, Hongkai Xiong, Qi~Tian, et~al.
\newblock Umg-clip: A unified multi-granularity vision generalist for open-world understanding.
\newblock In \emph{European Conference on Computer Vision}, pages 259--277. Springer, 2025.

\bibitem[Cui et~al.(2022)Cui, Zhou, Guo, Yin, Wu, Yoshie, and Chen]{cui2022contrastive}
Quan Cui, Boyan Zhou, Yu~Guo, Weidong Yin, Hao Wu, Osamu Yoshie, and Yubo Chen.
\newblock Contrastive vision-language pre-training with limited resources.
\newblock In \emph{European Conference on Computer Vision}, pages 236--253. Springer, 2022.

\bibitem[Fan et~al.(2024)Fan, Krishnan, Isola, Katabi, and Tian]{fan2024improving}
Lijie Fan, Dilip Krishnan, Phillip Isola, Dina Katabi, and Yonglong Tian.
\newblock Improving clip training with language rewrites.
\newblock \emph{Advances in Neural Information Processing Systems}, 36, 2024.

\bibitem[Zhang et~al.(2022)Zhang, Jiang, Miura, Manning, and Langlotz]{zhang2022contrastive}
Yuhao Zhang, Hang Jiang, Yasuhide Miura, Christopher~D Manning, and Curtis~P Langlotz.
\newblock Contrastive learning of medical visual representations from paired images and text.
\newblock In \emph{Machine Learning for Healthcare Conference}, pages 2--25. PMLR, 2022.

\bibitem[Huang et~al.(2021)Huang, Shen, Lungren, and Yeung]{huang2021gloria}
Shih-Cheng Huang, Liyue Shen, Matthew~P Lungren, and Serena Yeung.
\newblock Gloria: A multimodal global-local representation learning framework for label-efficient medical image recognition.
\newblock In \emph{Proceedings of the IEEE/CVF International Conference on Computer Vision}, pages 3942--3951, 2021.

\bibitem[Rizvi et~al.(2023)Rizvi, Tang, Jiang, Ma, and Hu]{rizvi2023local}
Syed~A Rizvi, Ruixiang Tang, Xiaoqian Jiang, Xiaotian Ma, and Xia Hu.
\newblock Local contrastive learning for medical image recognition.
\newblock In \emph{AMIA Annual Symposium Proceedings}, volume 2023, page 1236. American Medical Informatics Association, 2023.

\bibitem[Li et~al.(2024)Li, Yang, Ren, Nie, Gao, Tan, and Li]{li2024mlip}
Zhe Li, Laurence~T Yang, Bocheng Ren, Xin Nie, Zhangyang Gao, Cheng Tan, and Stan~Z Li.
\newblock Mlip: Enhancing medical visual representation with divergence encoder and knowledge-guided contrastive learning.
\newblock In \emph{Proceedings of the IEEE/CVF Conference on Computer Vision and Pattern Recognition}, pages 11704--11714, 2024.

\bibitem[Liu et~al.(2023)Liu, Lu, Wei, Wu, Wang, Zhang, and Zheng]{liu2023improving}
Bo~Liu, Donghuan Lu, Dong Wei, Xian Wu, Yan Wang, Yu~Zhang, and Yefeng Zheng.
\newblock Improving medical vision-language contrastive pretraining with semantics-aware triage.
\newblock \emph{IEEE Transactions on Medical Imaging}, 2023.

\bibitem[Chen et~al.(2023)Chen, He, Xue, Ge, Li, and Yang]{chen2023knowledge}
Xiaofei Chen, Yuting He, Cheng Xue, Rongjun Ge, Shuo Li, and Guanyu Yang.
\newblock Knowledge boosting: Rethinking medical contrastive vision-language pre-training.
\newblock In \emph{International Conference on Medical Image Computing and Computer-Assisted Intervention}, pages 405--415. Springer, 2023.

\bibitem[Liu et~al.(2024)Liu, Lu, and Wang]{liu2024towards}
Bo~Liu, Zexin Lu, and Yan Wang.
\newblock Towards medical vision-language contrastive pre-training via study-oriented semantic exploration.
\newblock In \emph{Proceedings of the 32nd ACM International Conference on Multimedia}, pages 4861--4870, 2024.

\bibitem[Wang and Wang(2023)]{wang2023umcl}
Yuhao Wang and Guangyu Wang.
\newblock Umcl: Unified medical image-text-label contrastive learning with continuous prompt.
\newblock In \emph{2023 IEEE International Conference on Bioinformatics and Biomedicine (BIBM)}, pages 2285--2289. IEEE, 2023.

\bibitem[Wu et~al.(2023)Wu, Zhang, Zhang, Wang, and Xie]{wu2023medklip}
Chaoyi Wu, Xiaoman Zhang, Ya~Zhang, Yanfeng Wang, and Weidi Xie.
\newblock Medklip: Medical knowledge enhanced language-image pre-training for x-ray diagnosis.
\newblock In \emph{Proceedings of the IEEE/CVF International Conference on Computer Vision}, pages 21372--21383, 2023.

\bibitem[Zhang et~al.(2023)Zhang, Wu, Zhang, Wang, and Xie]{zhang2023knowledge}
Xiaoman Zhang, Chaoyi Wu, Ya~Zhang, Yanfeng Wang, and Weidi Xie.
\newblock Knowledge-enhanced pre-training for auto-diagnosis of chest radiology images.
\newblock \emph{arXiv e-prints}, pages arXiv--2302, 2023.

\bibitem[Johnson et~al.(2019)Johnson, Pollard, Berkowitz, Greenbaum, Lungren, Deng, Mark, and Horng]{johnson2019mimic}
Alistair~EW Johnson, Tom~J Pollard, Seth~J Berkowitz, Nathaniel~R Greenbaum, Matthew~P Lungren, Chih-ying Deng, Roger~G Mark, and Steven Horng.
\newblock Mimic-cxr, a de-identified publicly available database of chest radiographs with free-text reports.
\newblock \emph{Scientific data}, 6\penalty0 (1):\penalty0 317, 2019.

\bibitem[Irvin et~al.(2019)Irvin, Rajpurkar, Ko, Yu, Ciurea-Ilcus, Chute, Marklund, Haghgoo, Ball, Shpanskaya, et~al.]{irvin2019chexpert}
Jeremy Irvin, Pranav Rajpurkar, Michael Ko, Yifan Yu, Silviana Ciurea-Ilcus, Chris Chute, Henrik Marklund, Behzad Haghgoo, Robyn Ball, Katie Shpanskaya, et~al.
\newblock Chexpert: A large chest radiograph dataset with uncertainty labels and expert comparison.
\newblock In \emph{Proceedings of the AAAI conference on artificial intelligence}, volume~33, pages 590--597, 2019.

\bibitem[Lei et~al.(2023)Lei, Dai, Jiang, Wu, Zhang, Zhang, Yao, Xie, Zhang, Li, et~al.]{lei2023unibrain}
Jiayu Lei, Lisong Dai, Haoyun Jiang, Chaoyi Wu, Xiaoman Zhang, Yao Zhang, Jiangchao Yao, Weidi Xie, Yanyong Zhang, Yuehua Li, et~al.
\newblock Unibrain: Universal brain mri diagnosis with hierarchical knowledge-enhanced pre-training.
\newblock \emph{arXiv preprint arXiv:2309.06828}, 2023.

\bibitem[Hinton(2015)]{hinton2015distilling}
Geoffrey Hinton.
\newblock Distilling the knowledge in a neural network.
\newblock \emph{arXiv preprint arXiv:1503.02531}, 2015.

\bibitem[Gao et~al.(2022)Gao, Liu, Xu, Zhang, Li, Ji, and Shen]{gao2022pyramidclip}
Yuting Gao, Jinfeng Liu, Zihan Xu, Jun Zhang, Ke~Li, Rongrong Ji, and Chunhua Shen.
\newblock Pyramidclip: Hierarchical feature alignment for vision-language model pretraining.
\newblock \emph{Advances in neural information processing systems}, 35:\penalty0 35959--35970, 2022.

\bibitem[Andonian et~al.(2022)Andonian, Chen, and Hamid]{andonian2022robust}
Alex Andonian, Shixing Chen, and Raffay Hamid.
\newblock Robust cross-modal representation learning with progressive self-distillation.
\newblock In \emph{Proceedings of the IEEE/CVF Conference on Computer Vision and Pattern Recognition}, pages 16430--16441, 2022.

\bibitem[Feng and Patras(2023)]{feng2023maskcon}
Chen Feng and Ioannis Patras.
\newblock Maskcon: Masked contrastive learning for coarse-labelled dataset.
\newblock In \emph{Proceedings of the IEEE/CVF Conference on Computer Vision and Pattern Recognition}, pages 19913--19922, 2023.

\bibitem[Yang et~al.(2024)Yang, Yang, Hui, Zheng, Yu, Zhou, Li, Li, Liu, Huang, et~al.]{yang2024qwen2}
An~Yang, Baosong Yang, Binyuan Hui, Bo~Zheng, Bowen Yu, Chang Zhou, Chengpeng Li, Chengyuan Li, Dayiheng Liu, Fei Huang, et~al.
\newblock Qwen2 technical report.
\newblock \emph{arXiv preprint arXiv:2407.10671}, 2024.

\bibitem[He et~al.(2016)He, Zhang, Ren, and Sun]{he2016deep}
Kaiming He, Xiangyu Zhang, Shaoqing Ren, and Jian Sun.
\newblock Deep residual learning for image recognition.
\newblock In \emph{Proceedings of the IEEE conference on computer vision and pattern recognition}, pages 770--778, 2016.

\bibitem[Du et~al.(2021)Du, Qian, Liu, Ding, Qiu, Yang, and Tang]{du2021glm}
Zhengxiao Du, Yujie Qian, Xiao Liu, Ming Ding, Jiezhong Qiu, Zhilin Yang, and Jie Tang.
\newblock Glm: General language model pretraining with autoregressive blank infilling.
\newblock \emph{arXiv preprint arXiv:2103.10360}, 2021.

\bibitem[Bakas et~al.(2018)Bakas, Reyes, Jakab, Bauer, Rempfler, Crimi, Shinohara, Berger, Ha, Rozycki, et~al.]{bakas2018identifying}
Spyridon Bakas, Mauricio Reyes, Andras Jakab, Stefan Bauer, Markus Rempfler, Alessandro Crimi, Russell~Takeshi Shinohara, Christoph Berger, Sung~Min Ha, Martin Rozycki, et~al.
\newblock Identifying the best machine learning algorithms for brain tumor segmentation, progression assessment, and overall survival prediction in the brats challenge.
\newblock \emph{arXiv preprint arXiv:1811.02629}, 2018.

\bibitem[Chatterjee et~al.(2022)Chatterjee, Nizamani, N{\"u}rnberger, and Speck]{chatterjee2022classification}
Soumick Chatterjee, Faraz~Ahmed Nizamani, Andreas N{\"u}rnberger, and Oliver Speck.
\newblock Classification of brain tumours in mr images using deep spatiospatial models.
\newblock \emph{Scientific Reports}, 12\penalty0 (1):\penalty0 1505, 2022.

\bibitem[Long et~al.(2015)Long, Shelhamer, and Darrell]{long2015fully}
Jonathan Long, Evan Shelhamer, and Trevor Darrell.
\newblock Fully convolutional networks for semantic segmentation.
\newblock In \emph{Proceedings of the IEEE conference on computer vision and pattern recognition}, pages 3431--3440, 2015.

\end{thebibliography}



\end{document}